# CONTOUR AND CENTRELINE TRACKING OF VESSELS FROM ANGIOGRAMS USING THE CLASSICAL IMAGE PROCESSING TECHNIQUES

BY

**IRINA ANDRA TACHE**∗

University Politehnica of Bucharest,
Department of Automatic Control and Systems Engineering

**Abstract.** This article deals with the problem of vessel edge and centerline detection using classical image processing techniques due to their simpleness and easiness to be implemented. The method is divided into four steps: the vessel enhancement which implies a non-linear filtering proposed by Frangi, the thresholding using Otsu method and the contour detection using the Canny edge detector due to its good performances for the small vessels and the morphological skeletonisation. The algorithms are tested on real data collected from a cardiac catheterism laboratory and it is accurate for images with good spatial resolution (512*512). The output image can be used for further processing in order to find the vessel length or its radius.

∗Corresponding author; *e-mail*: irina.tache@acse.pub.ro, irina.andra@gmail.com

# 1. Introduction

Angiography is a still a standard technique used to evaluate abnormalities occurred in blood vessels, providing images with good spatial and temporal resolutions.

The blood vessels enter in the category of soft tissues and have attenuation values close to water. Because blood cannot be distinguished from the surrounding soft tissue, a contrast agent is injected into the vessels in order to increase their visualization on a radiographic image. The contrast material used in angiography is an iodine based substance, because it has high atomic number and physical density and it is not toxic to the body, even if a large quantity is injected into the blood stream. The iodine concentration is adjusted in order to visually distinguish the blood vessels from their surroundings tissues.

A compromise between the quantity of the contrast agent administrated to the patient and the levels of radiation is made and this will limit the image resolutions.

The drawbacks of this clinical investigation procedure are: the invasiveness and the irradiation of the patient.

Usually, from this medical investigation the morphological information needs to be extracted.

Therefore, the processing of X-ray angiograms engaged a lot of scientists in the last decades in finding a quick and robust method to be implemented on medical equipment. A couple of algorithms are already available on angiographs for vessel contour detection and centerline.

The image quality is an important prerequisite and it is usually related to contrast which is judged as the difference in gray tones between the objects and their background. The better the contrast of the image is, the easier will be for the physician to assess the normal or the abnormal anatomical structures.

The vessels' analysis from an image consists of an examination of their topology or geometry, the gray levels and the neighborhood information.

The delineation of blood vessels on angiograms is an essential step in the computer aided diagnosis of the vascular diseases, although this is a challenging task due to the complexity of the image artifacts. Image noises have Poisson or Gaussian distribution.

The key step in improving the medical images for a quick diagnostic is the segmentation process which can be regarded as pixels classification into two groups: the object and the background. This step is necessary for centreline and contour detection of the vessels. This target is not properly achieved if the noise is not correctly filtered in the original image.

For the coronary angiograms when the heart and respiration movements are important, the digital subtraction angiography could not be applied. An alternative for the noise reduction is the multiscale filtering which have the advantage of optimally linking the spatial-frequency domains. They are used to eliminate the background structures and to enhance the vessels, especially the smaller ones. The resulted image is thresholded and further used by the centerline and edge detection algorithms.

The paper is organised as following: in the part 2 the spatial processing algorithms are detailed, in the part 3 the proposed algorithm is described and in the part 4, the results are analysed based on real data. Finally, the conclusions and further research directions are given.

## 2. Classical Image Processing Techniques

The main mathematical operations applied to one dimensional signals are adapted for the two dimensional case.

The main tasks of the computer scientists and engineers are the noise removal and the feature extractions from images. Therefore, some different filtering operations were developed each one dedicated to solve specific problems on certain image types.

### 2.1. Morphological Operations

The morphological mathematics was first developed by Georges Matheron in 60's. It is based on set theory is dedicated to the extraction of the object shape, such as the contours and the skeletons.

The morphological operators imply window operations applied on black-white or grayscale images. The set of all white pixels into an image is the morphological description of that image (Gonzalez and Woods, 2007).

The basic operations are: erosion (the suppression of the pixels next to the object's border), dilation (the addition of pixels next to the border), closing (the filling of small spaces between the objects' borders), skeletonisation (offers a compact representation of the object), pruning algorithm (the removal of unwanted lines). They work fine for images with uniform contrast and without important noises. Still, they are time consuming and a parallelization of the algorithms can be found in (Bräunl, 2001).

### 2.2. Angiogram Filtering

Even if the numerical imaging has revolutionized the medical image acquisition, there are still susceptible of noises induced by the electronic image transmission chain, by the sensor, by the electromagnetic fields nearby the acquisition equipment, etc.

The difficulty to analyze the X-ray angiograms comes from the low image contrast due to emitting a minimum dose of radiation to the patient. The contrast resolution for the coronary case, it has $2^8$ different gray levels. In addition to this, the high contrast anatomical structures such as bones can create shadows on the image and make the vessels investigation even harder.

In Alejandro *et al.* (1998) it is mentioned that an enhancement of the angiograms is an essential step for arterial problems diagnosis, angioplasty and bypass surgeries.

The noise reduction or the enhancement of certain components in the image is made by filtering operations. The smoothing operations include the median filtering which reduces the noise by replacing the value of each pixel, by its median gray value of its neighborhood.

Usually, the noises have a Poisson or Gaussian distribution.

Gaussian filters can improve the image, if the form and size of the filter are set accordingly to the standard deviation (s) of the Gaussian noise.

For example, a noise with a standard deviation of 0.5 will imply a mask size of 6*s which gives 5. So a 5x5 mask is used, with its coefficients computed as in Eq. (1):

$$g(x, y) = \frac{1}{2\pi s^2} e^{\frac{-\left((x-x_c)^2 + (y-y_c)^2\right)}{2s^2}} \qquad (1)$$

where $x_c$ and $y_c$ are the coordinates of the mask centre.

A small mask can cause less blurring to the filtered image and it permits the tagging of sharp lines in opposition to a large one which increase the blurring effect and it is used for large edges detection.

All these techniques are useful for removing the noise, but in return the object contour can be affected. Because the contour is critical in the image processing, the filtering methods must preserve it as much as possible.

The vessel enhancement techniques can be used for background components removal and for enhancing the small vessels, with low contrast. There are nonlinear filtering methods, which deal with the problem of non-additive noises without a normal distribution. They search in the images the geometrical structures which are the best candidates to represent the vessels.

Considering the vessels as connected, tubular and linearly piecewise curvilinear orientations structures, some filtering techniques are specifically designed (Sato *et al.*, 1998; Meijering *et al.*, 1999; Zwet *et al.*, 1995) for them. There are also other methods such as ridge detectors or morphological operations, but with more applicability in the retinal vessels. Due to the artifacts complexity found in the cardiac and cerebral X-ray imagistics, these algorithms are time consuming and fail to detect vessel area on low contrast images (Cao *et al.*, 2005).

Between the Gabor filtering and Frangivesselness filtering the second one is investigated into this paper due to its better performances in angiograms processing as it was concluded in (Tache, 2015). Both of them use the capacity of the Gaussian function to obtain iteratively the magnitudes and the directions of the vessel structure. In order to cope with the problem of vessels' diameter variation, a multivariate scale is included.

For avoiding the noise amplification, a prior median filtering can be done.

### 2.3. Skeletonisation

The topological skeletonisationis the image processing which frequently follows the segmentation step. It is a compact representation of the objects with line structures. Considering the object shape (F) and its maximal disk D which respects the inclusion from Eq. (2):

$$D \subseteq F \qquad (2)$$

The shape skeleton – $s(D)$ will be the set of maximal disks centers - $A_r(D)$ with r the shape radius (Petrou and GarcíaSevilla, 2006) and mathematically expressed as:

$$s(D) = \cup A_r(D) \qquad (3)$$

For example, the skeleton of the vessel segment is in fact its centerline, which is equidistant to both borders. It can be computed with morphological operations, but it is sensible to noises (Vertan, 1999). This happens partially due to the fact that these operations work fine only for images with a uniform contrast. Nevertheless, this can be corrected with the aid of some special designed structural elements (Petrou and GarcíaSevilla, 2006) and by pruning algorithm which is dedicated to remove the spurs (parasitic components) which are non-important in the overall shape of the object. The main idea is to remove all braches shorter than a pre-established number of pixels.

From the end points, a certain number of points (m) will be eliminated from each branch. A morphological dilatation is applied on the new end points with the aid of a structural element which depends on the number m, in order not to shorten the eventual main lines.

### 2.4. Centerline Extraction

This is an essential step in the 3D vessel reconstruction because it preserves its curvilinear shape and length.

The most important methods for accomplishing this operation are: the ridge detectors based on level set theory, morphological skeleton, contour pruning, fast matching and geometrical model based methods, filtering based methods and vessels' tracking. For improving the results, some morphological operations such as erosion or closing should be applied to the images before and/or after the centerline extraction.

Sang *et al*. (2004) has succeeded to extract the centerline from the cerebral blood vessels in poor contrast and noisy background, as for example in the cases when the digital subtraction angiography methods cannot be used. They used the Gabor filtering by choosing a large scale. Still, due to the complicated artifacts of the X-ray angiograms, it can fail to detect all the centerlines.

Another solution for the centreline detection is the morphological skeletonisation which exploits the vascular shape from where the connectivity map can be constructed.

The tracking methods use a model to track the vessels path starting from a seed point and choosing the following vessel candidate based on a set of attributes, such as, direction, width and center point. The algorithm inherits the control theory concepts adapted to digital imaging and a good review dedicated to biomedical applications can be found in (Acton and Ray, 2005).

An alternative of this algorithm could be the manual selection of the intermediary pixels of the centerline. A method was implemented in Matlab®, to interpolate these points by a cubic spline function. The model comprises piecewise polynomialsof third-orderwhich must verify separately a set of n control points, given by the user. A simpletridiagonal system of $n-2$ equations is formed by choosing the boundary conditions from the zero setting of the secondorder derivativeof eachpolynomial at their endpoints (Bartels *et al*., 1995).

### 2.5. Edge Detection

The edge or contour detection is an important image processing which can be regarded as a segmentation step. It is defined as the linear characteristic of the object which has at least one neighbour outside the object. Actually, it implies the detection of the discontinuities in the gray levels.

Canny (1986) proposed a method for image gradient computation in the neighbourhood pixels and its steps are presented in the followings. The Gaussian filtered image which is a slightly blurred version of the original image is computed. Two Sobel operators can approximate the image gradient and a searching for the local maximum admitted by the output magnitudes in the gradient direction is performed. For limiting the multiple maximum detections due to the existing noises, a hysteresis thresholding is applied for detecting only the strong edges and their low connected branches using the direction information. This algorithm is more robust to noises and can detect the low contrast edges.

## 3. The Proposed Algorithm

Firstly, the vessels are enhanced using two consecutive filtering methods:
1. Median filtering - for avoiding the noise amplification
2. Frangivesselness filtering - detects the pixels which are the best candidates to represent the vessel.
3. After processing the maximum magnitude response, the edge and centerline are detected and some further tracking algorithms can be applied.
4. The graph nodes are defined at image coordinates where the probabilities of vessel occurrence are high.
5. The shortest paths from an initial point to all nodes in the graph are found by the Dijkstra's method.

The algorithm steps are presented in the Fig 1.

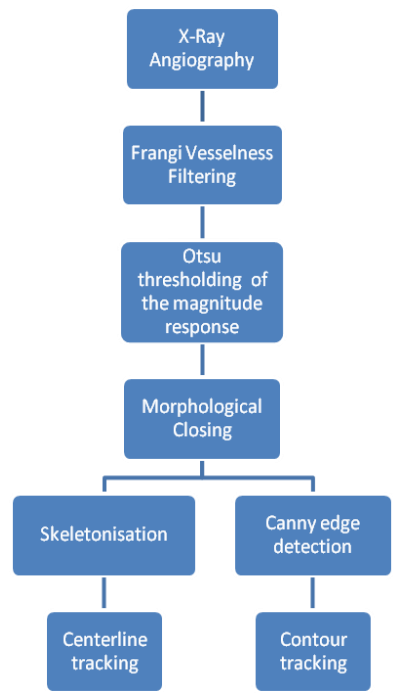

Fig. 1 – Image Processing Algorithm.

## 4. Results

The clinical protocol includes the acquisition of the patient fluoroscopic angiograms from different view angles and the EKG signal associated to each image set. The spatial resolution is 512*512 pixels and the temporal resolution is around 15 images/second.

Images are stored in DICOM format, which is a set of standards for handling, storing and transmitting information in medical images. The file contains data about the acquisition setting of the medical equipment such as the rotation angles, the distances between the main components of the angiograph, the frames rate and the numerical image/images.

The coronary angiograms from 10 patients with the visualisation of the initial bolus in the first cardiac cycles were selected for study.

Some image samples are provided in Table 1 for evidentiate the progressively mixing of the contrast agent with the blood and its evolution throughout the cardiac cycle.

**Table 1**
*A Sample of Coronarography Projections*

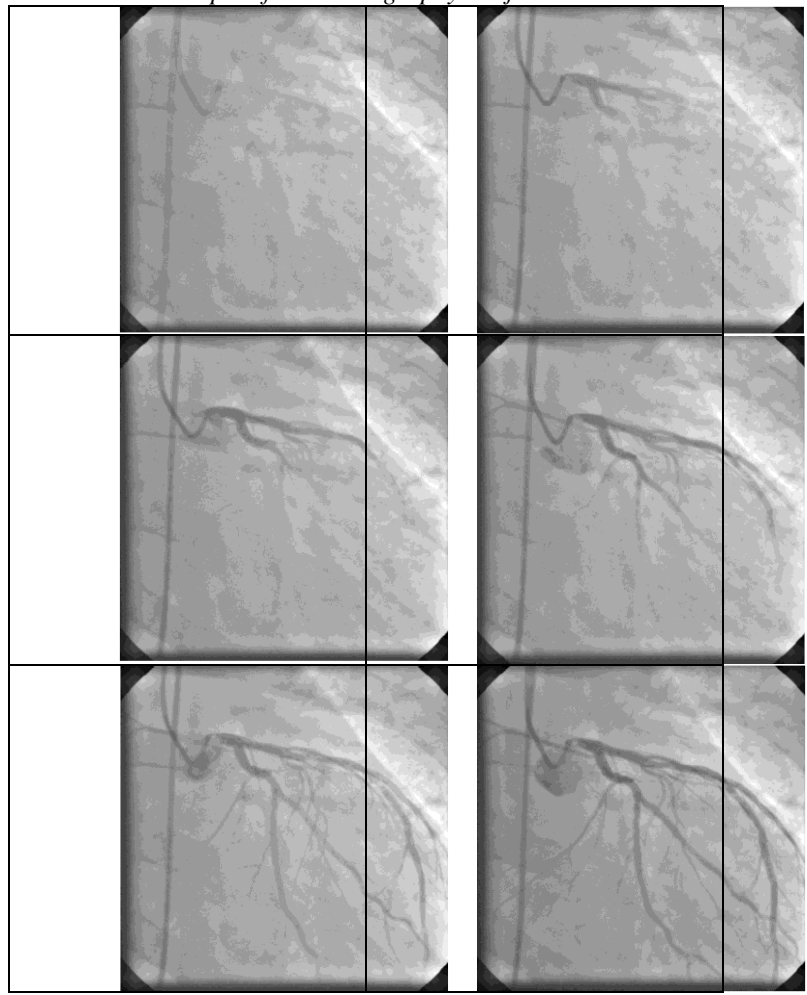

The movement artefacts must be avoided by choosing the images from the final diastole phase where the coronary vessels tend to remain motionless.

The image processing algorithm described in Fig. 1 was implemented in Matlab® and tested on a regular laptop with a medium frequency processor (around 2.6 GHz) with a minimum of 4 GB RAM.

The output images of the Frangivesselness for a cardiac angiogram and their processing (thresholding, noise removal with morphological operations, skeletonisation and edge detection) are revealed in the Table 2.

In the Step 1 it is revealed a colour map with the direction information resulted from the Frangivesselness filtering. In the Step 2, the magnitude response is shown and it is used for further processing.

**Table 2**
*Edge Detection and Skeletonisation*

| 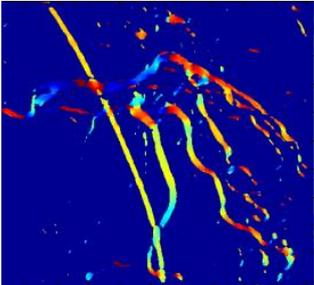 | 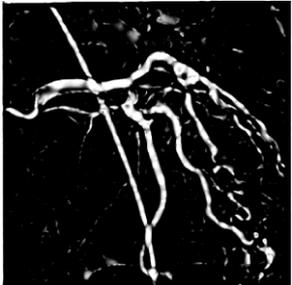 |
|---|---|
| 1. Image with direction information | 2. The magnitude response |
| 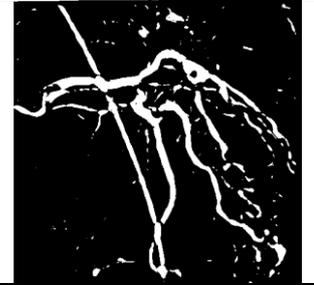 | 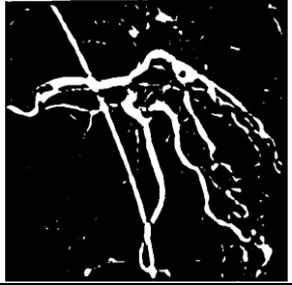 |
| 3. Otsu thresholding | 4. Morphological closing |
| 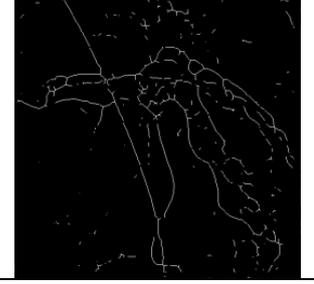 | 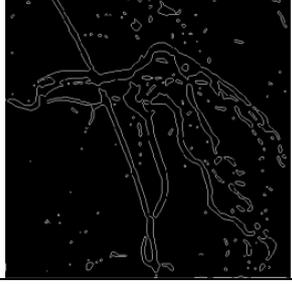 |
| 5. Skeletonisation | 6. Canny edge detection |

Thirdly, the Otsu'sglobalthresholding is applied for obtaining a segmentation of the vessels. In the Step 4, a closing morphological operation is applied for filling the small gaps of the objects' borders.

In the fifth step, the morphological skeletonisation is applied and in the sixth step the Canny algorithm is used for finding the vessels edges. In both cases there are residual spurs which must be eliminated with pruning operations.

Also, these results can be used for finding the graph nodes in order to construct the adjacency matrix of the pixels by searching the connecting components in the neighbourhood of the pixels. For defining the local neighbourhood connection, on the images from the steps 5 and 6 a filtering operation is applied using a window size of [5 5].

On the resulted images the minimal path cost of the nodes which are orthogonal to the tracking direction must be determined. This is accomplished with the well-known Dijkstra's algorithm by using specific cost functions, such as, the distance between the nodes, their position, the information about the gray levels intensities and orientations extracted from the vesselness filtering. By minimizing the costs, a new path between the nodes/pixels is created and the edges are drawn in order to reconstruct the curvilinear shapes of the vessel.

Still, this method would not be completely independent of the human intervention, because it may imply the user selection of a start point on the vessel structure in order to construct the connectivity matrix. For relevant results, the above mentioned method requires images with good contrast.

## 5. Conclusions

For the coronary angiograms when the heart and the respiration movements are important, the digital subtraction angiography could not be applied. An alternative for the noise reduction is the multiscale filtering which have the advantage of optimally linking the spatial-frequency domains. They are used to eliminate the background structures and to enhance the vessels, especially the smaller ones. The resulted image is globalthresholded and further used for centerline and edge detection.

The proposed algorithms are applied for real clinical data and their limitations are evidenced in the edge detection of vessels with high curvature.

The main goal of this spatial processing is to serve as an input for the minimal costs Dijkstra's algorithm.

Sparse nodes are achieved by finding local maximum of Frangivesselness map.

The method works well for all image sets and the vessel sizes were computed. Still the performance degrades when there is background clutter in the angiogram, such as spines, catheters, and guidewires.

Furthermore, vessel crossings could be incorrectly identified as vessel bifurcations.

The results can be used for measuring the 2D length and radius of a vessel's segment selected by an image expert. Also, the centreline and the edges of the vessel segment can be used for the three dimensional reconstruction from two monoplane angiography.

Further work includes the refinement of the vessel structure extracted from a 2D angiogram by using prior shape information from a 3D coronary vessel model of the same patient.

**Acknowledgments.** This research was funded by the UEFISCDI National Project II for Partnership, Grant No. 130/29.07.2012 "High PErformance Computing of PersonAlizedCaRdioComponenT Models".

**REFERENCES**

Acton S.T., Ray N., *Biomedical Image Analysis: Tracking*, Morgan & Claypool Publishers (2005).

Alejandro F.F., Niessen W.J., Vincken K.L., Viergever M.A., *Multiscale Vessel Enhacement Filtering*, Springer Verlag, pp. 130-137 (1998).

Bartels R.H., Beatty J.C., Barsky B.A., *An Introduction to Splines for Use in Computer Graphics and Geometric Modelling*, The Morgan Kaufmann Series in Computer Graphics, San Francisco, pp. 9-17 (1995).

Bräunl T., *Tutorial in Data Parallel Image Processing*, Australian Journal of Intelligent Information Processing Systems (AJIIPS), vol. 6, no. 3, pp. 164-174 (11) (2001).

Canny J., *A Computational Approach to Edge Detection*, IEEE Trans. Pattern Analysis and Machine Intelligence, vol. 8, no. 6, pp. 679-698 (1986).

Cao Z., Liu X., Peng B., Moon Y.S., *DSA Image Registration based on Multiscale Gabor Filters and Mutual Information*, IEEE International Conference on Information Acquisition, Hong Kong and Macau, China (2005).

Gonzalez R., Woods R., *Digital Image Processing*, 3rd Edition, Chapter 9, Pearson, Prentice Hall (2007).

Meijering E.H., Zuiderveld K.J., Viergever M.A., *Image Registration for Digital Subtraction Angiography*, Int. J. Comput. Vis., vol. 31, pp. 227-246 (1999).

Petrou Maria, García Sevilla Pedro, *Image Processing Dealing with Texture*, John Wiley & Sons Publishing (2006).

Sang N., Tang Q., Liu X., Weng W., *Multiscale Centerline Extraction of Angiogram Vessels Using Gabor Filters*, First International Symposium, Computational and Information Science, Shanghai, China, pp. 570-575 (2004).

Sato Y., Nakajima S., Shiraga N., Atsumi H., Yoshida S., Koller T., Gerig G., Kikinis R., *3D Multi-Scale Line Filter for Segmentation and Visualization of Curvilinear Structures in Medical Images*, Medical Image Analysis, vol. 2, no. 2, pp. 143-168 (1998).

Tache I.A., *Vessels Enhancement in X-Ray Angiograms*, IEEE International Conference on e-Health and Bioengineering, Iași, 2015, DOI: 10.1109/EHB.2015.7391549.

van der Zwet P.M., Reiber J.H., *The Influence of Image Enhancement and Reconstruction on Quantitative Coronary Arteriography*, Int. J. Card. Imaging, vol. 11, pp. 211-21 (1995).
Vertan C., *Prelucrarea şi Analiza Imaginilor*, Printech Ed., Bucureşti (1999).

# DETECTAREA CONTURULUI ŞI A LINIEI CENTRALE ALE VASELOR DE SÂNGE DIN AGIOGRAFII FOLOSIND TEHNICILE CLASICE DE PROCESARE A IMAGINILOR

(Rezumat)

Acest articol tratează problema detecţiei conturului şi liniei centrale ale vasului folosind tehnici clasice de procesare a imaginii datorită simplităţii lor şi a vitezei de calcul. Metoda este împărţită în cinci etape :

1. Se filtrează imaginea cu un filtru median şi apoi cu un filtru neliniar de tip Frangi, care va determina pixelii cu probabilitatea cea mai mare de a reprezenta vasele de sânge.

2. Se aplică operaţia morfologică de închidere.

3. Se binarizează imaginea folosind metoda de prag global a lui Otsu.

4. Fie se detectează conturul cu ajutorul detectorului de margini al lui Canny datorită bunelor sale performanţe pentru vasele mici, fie se aplică operaţia morfologică de scheletizare.

5. Se detectează nodurile grafului din imaginea de la pasul 4 şi se aplică algoritmul de calcul al costului minim al lui Dijkstra.

Imaginea rezultată poate fi utilizată pentru o procesare ulterioară cu scopul găsirii lungimii şi razei unui segment de vas selectat de un expert în imagini medicale.